\documentclass[a4paper,twoside]{article}

\usepackage{epsfig}
\usepackage{subcaption}
\usepackage{calc}
\usepackage{amssymb}
\usepackage{amstext}
\usepackage{amsmath}
\usepackage{amsthm}
\usepackage{multicol}
\usepackage{pslatex}
\usepackage{apalike}
\usepackage[bottom]{footmisc}
\usepackage{booktabs}
\usepackage{SCITEPRESS}
\begin{document}

\title{Finger-UNet: A U-Net based Multi-Task Architecture for Deep Fingerprint Enhancement}

\author{\authorname{Ekta Gavas\sup{1}\orcidAuthor{0000-0001-6437-3357}, Anoop Namboodiri\sup{1}\orcidAuthor{0000-0002-4638-0833} }
\affiliation{\sup{1}Center for Visual Information Technology, International Institute of Information Technology, Hyderabad, India}
\email{ekta.gavas@research.iiit.ac.in, anoop@iiit.ac.in}
}

\keywords{Fingerprint Enhancement, Fingerprint Quality, Image Enhancement, Multi-task learning}

\abstract{For decades, fingerprint recognition has been prevalent for security, forensics, and other biometric applications. However, the availability of good-quality fingerprints is challenging, making recognition difficult. Fingerprint images might be degraded with a poor ridge structure and noisy or less contrasting backgrounds. Hence, fingerprint enhancement plays a vital role in the early stages of the fingerprint recognition/verification pipeline. In this paper, we investigate and improvise the encoder-decoder style architecture and suggest intuitive modifications to U-Net to enhance low-quality fingerprints effectively. 
We investigate the use of Discrete Wavelet Transform (DWT) for fingerprint enhancement and use a wavelet attention module instead of max pooling which proves advantageous for our task. Moreover, we replace regular convolutions with depthwise separable convolutions, which significantly reduces the memory footprint of the model without degrading the performance. We also demonstrate that incorporating domain knowledge with fingerprint minutiae prediction task can improve fingerprint reconstruction through multi-task learning. Furthermore, we also integrate the orientation estimation task to propagate the knowledge of ridge orientations to enhance the performance further.
We present the experimental results and evaluate our model on FVC 2002 and NIST SD302 databases to show the effectiveness of our approach compared to previous works.}

\onecolumn
\maketitle
\normalsize 
\vfill

\section{\uppercase{Introduction}}
\label{sec:introduction}
Fingerprints are one of the most crucial biometric traits due to their characteristic of being unique and permanent (lifelong) to every individual \cite{jain2004introduction}. In addition, they are comparatively easy to acquire \cite{jain2004introduction}. So, fingerprints are very commonly used biometrics for identification. Also, the recognition of fingerprints collected from crime scenes is helpful in criminal investigations and forensic applications. However, the acquisition of 'good' quality fingerprints is not trivial. A fingerprint is identified based on its unique ridge-valley structure with a well-defined frequency, orientation, and the location of special points called minutia. In reality, a fingerprint may be degraded due to various reasons like sensor defects, noise, oily skin, finger cuts/wounds, or uneven pressure while acquiring. It may have poor ridge structure, overlapping backgrounds, and low contrast due to collection from crime scenes. Due to these factors, the performance of fingerprint systems gets affected. Fingerprint enhancement deals with this issue by uplifting the quality of fingerprints to, at the least, recover the fingerprint structure to the maximum extent possible. Whether it is matching or recognition, fingerprint enhancement thus becomes an essential step in pre-processing in cases where good-quality prints are rare.

Traditionally, filtering techniques and other classical image processing methods were used to improve the ridge clarity, and fingerprint quality \cite{chikkerur2005fingerprint,greenberg2002fingerprint,hong1998fingerprint,kim2002new,yang2002improved}. Then, with the advent of deep learning, neural networks, particularly convolutional neural networks (CNNs), are being employed to tackle this problem \cite{li2018deep,qian2019latent,joshi2019latent}. In this paper, to recover good quality fingerprints, we explore a popular architecture, U-Net \cite{ronneberger2015u}, essentially an encoder-decoder-style architecture with skip connections. 
This paper focuses on improving the basic U-Net architecture in order to improve fingerprint quality in a robust and intuitive manner. We suggest that these changes improve the fingerprint quality in addition to reducing the network parameters. 

Research studies show that CNNs, not being noise-robust, noise gets enlarged as data propagates through the layers of CNNs after several epochs of training which may significantly impact learning and can even lead to overfitting \cite{xie2019feature,geirhos2018imagenet,zhao2022wavelet,li2021wavecnet,liu2019multi}. The downsampling operation of CNNs is responsible for weak noise-robustness and loss of information. In the frequency domain, DWT can provide high-quality downsampling, significantly reducing this information loss. It decomposes the 2D image into four frequency components, and the noise-containing component is filtered and dropped; hence it is not forwarded into the network layers, avoiding noise propagation. In this paper, we use a Wavelet-Attention block proposed in \cite{zhao2022wavelet} in our U-Net architecture as a downsampling layer, which constrains the noise from high-frequency components obtained with DWT to propagate further, whereas information in low-frequency components is unaffected.

Further, we incorporate domain knowledge into our learning for the model to better understand the fingerprint structure. We achieve this by adding minutia-prediction and orientation estimation branches with a multi-task learning approach, which further helps to improve performance. Moreover, we replace standard convolution layers with depthwise separable convolutions \cite{chollet2017xception}, which does not impact the performance considerably but helps to reduce the model parameters drastically.

\subsection{Related Work}
In past decades, several works have been proposed for fingerprint enhancement that involves applying traditional or classical image processing techniques. The research focused on improving the ridge structure and increasing the contrast in images \cite{greenberg2002fingerprint}. It started with most basic techniques like histogram equalization \cite{ezhilmaran2014review} to using Fourier transforms \cite{sherlock1992algorithm,chikkerur2005fingerprint,rahman2008improved} to remove noise from images. \cite{chikkerur2005fingerprint} extended the use of short-term Fourier Transform (STFT) analysis to 2D fingerprint images and estimated the intrinsic properties of fingerprints. Gabor filtering was used to estimate the orientation and frequency fields which were later used to enhance fingerprints \cite{hong1998fingerprint,kim2002new,yang2002improved}. \cite{liu2014latent} proposed a dictionary-based approach where dictionaries were created with a set of Gabor filters, and the multi-scale representation is iteratively applied to recover the enhanced image. \cite{feng2012orientation} proposed a path-based dictionary approach for orientation estimation.

For many years, much research has been carried out on neural networks, especially CNNs. With this, many past works proposed deep architectures for fingerprint enhancement.  \cite{qian2019latent} introduced a deep network with dense blocks called DenseUNET to improve image quality in pixel-to-pixel and end-to-end manner. \cite{joshi2019latent} incorporated adversarial training using GANs for fingerprint enhancement. In another interesting work, \cite{adiga2019fpd} posed the fingerprint denoising problem as a segmentation (task) using M-net-based architecture. Few previous works had considered adding orientation knowledge to guide the enhancement task. \cite{li2018deep} proposed a deep architecture called FingerNet with enhancement and orientation deconvolution branches to enhance images using multi-task learning in two-stage training. They posed coarse orientation estimation as a classification problem with a quantized orientation field. 

Generally, CNNs work with the matrix of pixels in the spatial domain. In contrast, the frequency domain deals with how these pixel values change in the spatial domain. Several mathematical transforms exist in the frequency domain, including Fourier, Laplace, Z, and wavelet transform. Wavelet transform is widely used in signal processing applications, image denoising \cite{kimlyk2018image,ismael2016new}, and compression \cite{chowdhury2012image,kanagaraj2020image}. It decomposes the image information into signal details and approximations, commonly known as high-frequency and low-frequency components. Previous studies show several attempts have been made to incorporate DWT into CNNs \cite{bae2017beyond,liu2019multi,li2021wavecnet,zhao2022wavelet}. \cite{bae2017beyond} showed that CNNs could benefit from learning about wavelet subbands and proposed a wavelet residual network (WavResNet). 
\cite{duan2017sar} applied dual-tree complex wavelet transform (DT-CWT) and designed a Convolutional-Wavelet Neural Network (CWNN) to suppress noise and extract features robustly from SAR images. \cite{liu2019multi} proposed a multi-level wavelet CNN (MWCNN) model which integrates wavelet transform into CNN to reduce feature map resolution and increase receptive field. \cite{li2021wavecnet} designed DWT/IDWT layer for integration into deep networks. Later, \cite{zhao2022wavelet} modified these layers to include a wavelet attention module to retain detailed information in high-frequency components in DWT. In the domain of fingerprint biometrics also, research has been carried out to show the effectiveness of wavelet transform for fingerprint enhancement task \cite{zhang2002wavelet,hsieh2003effective}. Hence, this paper combines the advantages of deep networks over traditional filtering techniques and wavelet attention module to design an architecture for fingerprint enhancement.
\begin{figure*}
  \centering \includegraphics[height=6cm]{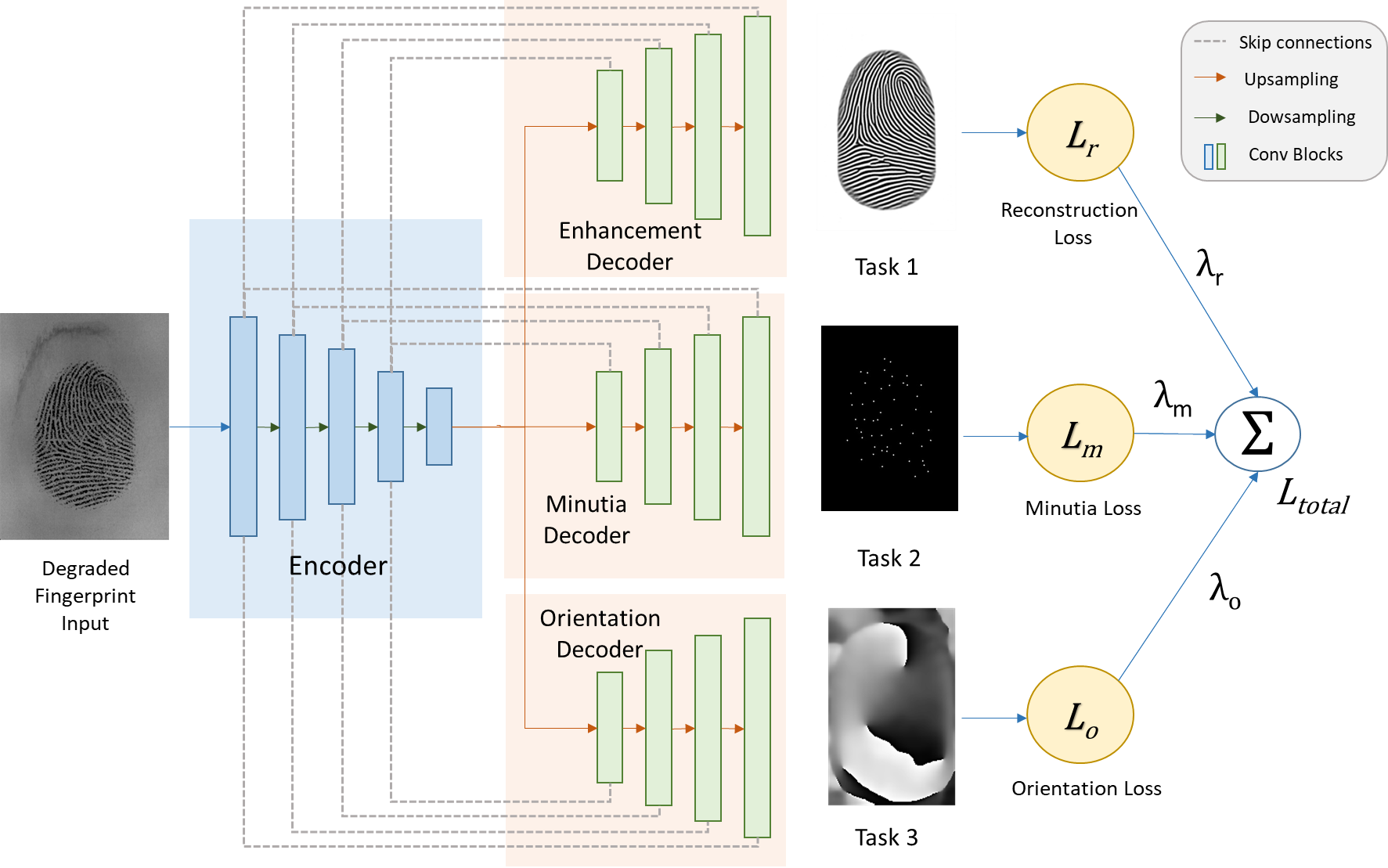}
  \caption{Architecture of our proposed approach with enhancement, minutia detection and orientation estimation branches in a multi-task learning setting}
  \label{fig:arch}
\end{figure*}
\subsection{Contributions}
We made the following contributions to this paper:
\begin{enumerate}
    \item We propose a U-Net-based architecture with the DWT-based wavelet attention block for fingerprint enhancement.
    \item We demonstrate that using minutia detection and orientation estimation branches in a multi-task manner can guide enhancement task to use domain knowledge to improve performance further.
    \item We evaluate our model with publicly available datasets FVC 2002 and NIST SD302, and show that our model performs well in the context of structural similarity, fingerprint quality, and matching.
\end{enumerate}
\section{\uppercase{Methodology}} 
The fingerprint enhancement task can be related to image denoising. However, we need to consider the inherent properties of biometric data and that it should be handled differently than normal real-world images. Autoencoder-style architectures are pretty popular for image-denoising tasks. We design an architecture called FingerUNET, based on the popular U-Net \cite{ronneberger2015u} originally proposed for biomedical image segmentation tasks but has been shown to perform well on fingerprint enhancement task \cite{qian2019latent,liu2020automatic}. In this paper, we offer a modified U-Net targeted to handle fingerprint data for enhancement task.

\subsection{Wavelet Transform as Pooling}
Wavelet transform can be said as a transformation that maps the signal to a multi-resolution representation. Wavelet has been combined with neural network for function approximation \cite{zhang1992wavelet}, signal representation and classification \cite{szu1992neural}. \cite{li2021wavecnet} proposed DWT layers that decompose a 2D image into its frequency components. Here, the pooling is performed with down-sampling operation instead of max pooling or average pooling to avoid information loss and aliasing effect. It also increases the noise robustness of CNNs.

Given 2D data $X$, the DWT usually does 1D DWT on every row and column, resulting in four frequency components $X_{LL}$, $X_{LH}$, $X_{HL}$, and $X_{HH}$. $X_{LL}$ is the low-frequency component of input X, representing the main information, including the basic structure in the image; $X_{LH}$, $X_{HL}$, and $X_{HH}$ are three high-frequency components that save the horizontal, vertical, and diagonal details of $X$, respectively. \cite{li2021wavecnet} designed these DWT/IDWT layers in Pytorch and made DWT/IDWT operations differentiable and compatible with CNNs.
\begin{equation}
\begin{split}
    X_{LL} = LXL^T, X_{LH} = HXL^T,
    \\
    X_{HL} = HXL^T, X_{HH} = HXH^T,
\end{split}
\end{equation}
where matrix $L$ and $H$ are the cyclic matrix composed of wavelet low-pass filter $\{l_k\}_{k\in Z}$ and high-pass filter $\{h_k\}_{k\in Z}$ respectively. Here $L$ and $H$ are as in \cite{li2021wavecnet}.
In FingerUNET, we make use of wavelet attention block (WA Block) proposed in \cite{zhao2022wavelet} built by modifying the above DWT layer. The wavelet attention block can be defined as
\begin{align}
    x_g &= \sigma (f (X_{LH}, X_{HL})) \\
    A_m &= X_{LL} * x_g \\
    Z &= f (X_{LL}, A_m)
\end{align}
where $f$ represents the feature aggregation function, $\sigma$ denotes the softmax function.
Once the four components are obtained using DWT, the WA block takes the horizontal feature $X_{LH}$ and vertical feature $X_{HL}$ and aggregates them by element-wise addition as a global detail feature. Then this feature is normalized using the softmax function. The normalized feature $x_g$ and the low-frequency component $X_{LL}$ are used to generate the attention map $A_m$ through element-wise multiplication. Finally, the original low-frequency component $X_{LL}$ is added to the attention map $A_m$ by element-wise operation and given as output $Z$. As the component $X_{HH}$ does not contain additional information for fingerprint ridges, it is not used in WA block.
We replace the max-pooling layers in vanilla U-Net with this WA block and notice an improvement in the fingerprint enhancement task. IDWT layer is used to reconstruct the output back to spatial domain \cite{li2021wavecnet}.

\subsection{Reconstruction Loss}
\label{sec:2.2}
The choice of the loss function is crucial in any neural network training. Earlier works \cite{burger2012image,dong2014learning} used mean squared error (MSE) or $l_2$ loss for reconstruction in image enhancement tasks. \cite{zhao2016loss} pointed out several limitations to using $l_2$ for image restoration tasks. $l_2$ does not correlate well with human perception of image quality \cite{zhang2012comprehensive}, due to the assumption that the impact of noise is independent of the local characteristics of the image. Moreover, $l_2$ penalizes larger errors but tolerates small ones without considering the underlying structure of the image. In contrast, the human visual system (HVS) is more sensitive to luminance and color variations in texture-less regions \cite{zhao2016loss,winkler2004visibility}. \cite{zhao2016loss} suggested using $l_1$ loss instead of $l_2$ as it does not over-penalize large errors and it also helps reduce artifacts introduced by $l_2$. We utilize the effectiveness of using $l_1$ as a reconstruction loss for fingerprint enhancement. Even though the images are grayscale, we witness a performance improvement compared to the $l_2$ counterpart, as demonstrated in later sections.

\subsection{Minutia Detection and Orientation Estimation}
Every fingerprint has a well-defined ridge-valley structure and a frequency associated with it. The presence of specific patterns like whorl, loop, and arch in a fingerprint at different locations makes it unique. Ridge endings and bifurcations (minutiae) are important in fingerprint matching. Hence, fingerprint images prove to be very different from normal real-world images and hence they need to be dealt with differently. These properties of fingerprint data should not be neglected in learning. For this, adding domain knowledge to our network training becomes useful to model the data better. Moreover, in the case of low-quality images where the structure is not intuitive, the knowledge of fingerprint properties can help the network to learn and predict the structure better. 

Previous works utilize ridge orientations when enhancing fingerprints \cite{hong1998fingerprint,kim2002new,yang2002improved,li2018deep}. Orientation estimation gives the direction of the gradient of the fingerprint segment which can be important in fingerprint enhancement. \cite{li2018deep} trained enhancement and orientation estimation branches in an end-to-end manner and demonstrated the effectiveness of orientation knowledge to aid the enhancement. The orientation estimation branch in our work predicts vectorized orientation fields, specifically sine and cosine values. This eliminates the need to segregate orientation patches into 20 fixed sets of classes \cite{li2018deep} and allows for the estimation of precise ridge orientation.

We also incorporate the minutia detection branch for using domain knowledge. \cite{darlow2017fingerprint} proposed a minutia extraction network called MENet, which outputs the minutiae probabilities map, which is later processed to determine precise minutia locations. To propagate the information about fingerprint properties to the enhancement branch, we use a gray-scale image with minutiae locations marked with a white dot on a black background as ground-truth as shown in the minutia branch (Task 2) image in Figure \ref{fig:arch}. 

We approach the problem at hand using multi-task learning by optimizing fingerprint enhancement, minutia detection, and orientation tasks simultaneously. We use a shared encoder that learns the feature representations from noisy fingerprint images and one decoder for each of the three tasks, as shown in Figure \ref{fig:arch}. We use the $l_1$ loss mentioned in Section \ref{sec:2.2} for the enhancement task denoted by $L_r$.  As we are solving orientation estimation as a regression problem, we use $l_2$ loss or MSE for this branch. We denote this loss by $L_o$. For the minutia detection branch, the values in the ground-truth map are zero or one based on the presence of minutia locations, hence we use binary cross entropy as the loss function and it is denoted by $L_m$.
The final loss \textit{L} is the summation of $L_r$, $L_m$, and  $L_o$, weighted by scalars $\lambda_r, \lambda_m$ and $\lambda_o$ respectively.
 
 \begin{equation}
 \label{loss_equation}
     L_{total} = \lambda_rL_r + \lambda_mL_m + \lambda_oL_o
 \end{equation}

\subsection{Operation Approximation with Depthwise Separable Convolution}
Standard convolution operation applies spatial and channel interactions by multiplying values over several spatial pixels and all the channels. The idea of depthwise separable convolution \cite{chollet2017xception} is to disentangle the two by using each filter channel only at one input channel. Then, we use a 1x1 filter to cover the depth dimension. Though this method is an approximation to the usual convolution operation, it does not incur a considerable drop in performance. In addition, it reduces the number of model parameters by 30\% in our experiments which in turn helps to avoid the over-fitting problem. Hence, in our work, we replace the convolution layers with depth-wise separable convolutions.

\section{\uppercase{Experiments}}
\subsection{Dataset}
\begin{figure}
  \centering \includegraphics[height=5cm, ]{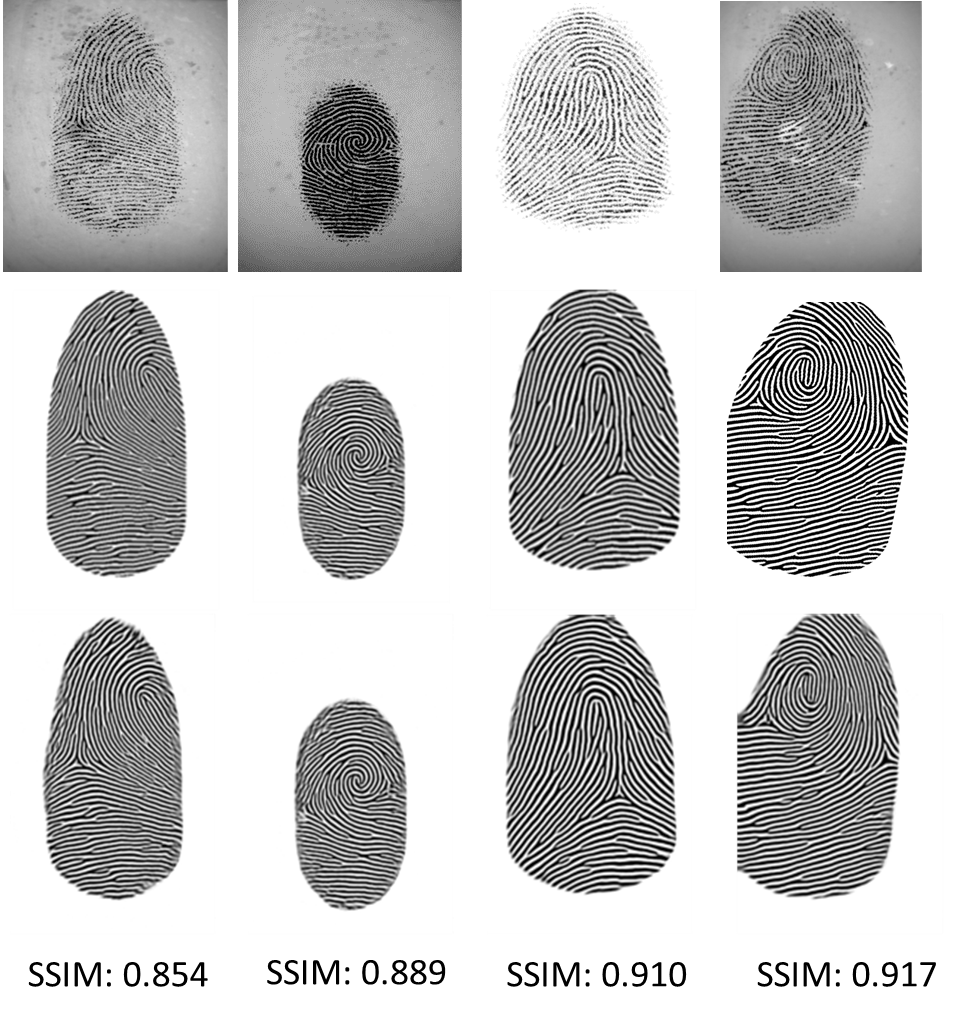}
  \caption{Enhanced images (bottom row) with corresponding degraded input (top row) and ground-truth (middle row) with our approach on SFinGe synthetic test set. SSIM is reported between each enhanced and ground-truth pair}
  \label{fig:ssim}
\end{figure}

The fingerprint datasets available in the public domain are either not huge enough to train a deep network or do not contain enhanced/clean impressions for ground-truth. So, for all the experiments in this paper, we are using synthetically generated fingerprints from SFinGe \cite{cappelli2002synthetic}. SFinGe can generate fingerprints with the required number of impressions and varying ridge structures and patterns. A wide range of noise and degradation like skin elasticity, noise, pressure, scratches, etc., can also be added. We generated 10,000 fingerprint pairs (degraded and ground-truth image pairs) for training, 1,000 for validation, and 3,000 for testing. The generated data contain varying types and degrees of available noises and backgrounds (optical, capacitive sensors, and no background). Figure \ref{fig:ssim} shows sample input and ground-truth pairs from the dataset in the first two rows. For evaluation, we use publicly available datasets FVC 2002 \cite{maio2002fvc2002} and NIST Special Database (SD) 302 \cite{fiumara2019nist}. FVC 2002 consists of fingerprints from optical, capacitive sensors, and synthetic fingerprints available in four sets, of which 80 images are publicly available in each set. NIST SD302 has plain, rolled and touch-free impressions captured from various devices. We use the subset 302d containing 5141 fingerprint images acquired from 4 different auxiliary devices.

\subsection{Training}
We use the PyTorch framework for all experiments in this paper. We apply various data augmentations like random translation, rotation, flip (horizontal/vertical), and shear, for the network to generalize well.  We apply the same degree of augmentations to each input-ground truth pair for consistency. The images are resized to fixed dimensions of 400x256. The hyperparameters are chosen using grid search. The model is trained with Adam optimizer with a learning rate of 0.001. The batch size is set as 32. The loss weights $\lambda_r, \lambda_m$, and $\lambda_o$ are 0.8, 0.1, 0.1 respectively. The network is trained on four GPUs in data parallel mode, and each GPU is NVIDIA GeForce TITAN X with 16 GB RAM. As the training is end-to-end, the losses from minutia and orientation guide the enhancement branch for better fingerprint output.

\begin{figure*}
\centering
\begin{subfigure}{.5\textwidth}
  \centering
  \includegraphics[width=0.9\textwidth]{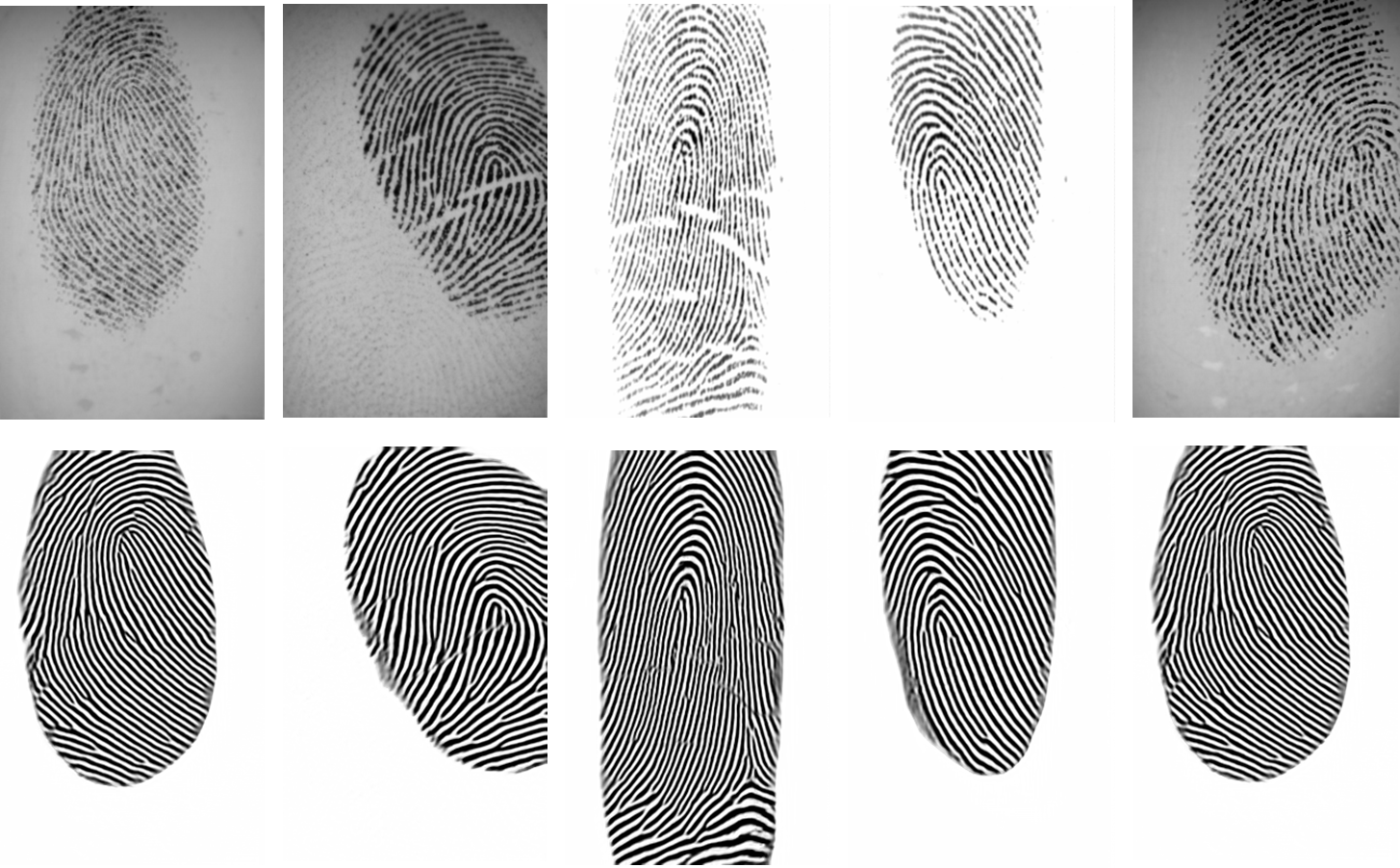}
  \caption{FVC 2002 dataset}
  \label{fig:fvc_results}
\end{subfigure}%
\begin{subfigure}{.5\textwidth}
  \centering
  \includegraphics[width=0.9\textwidth]{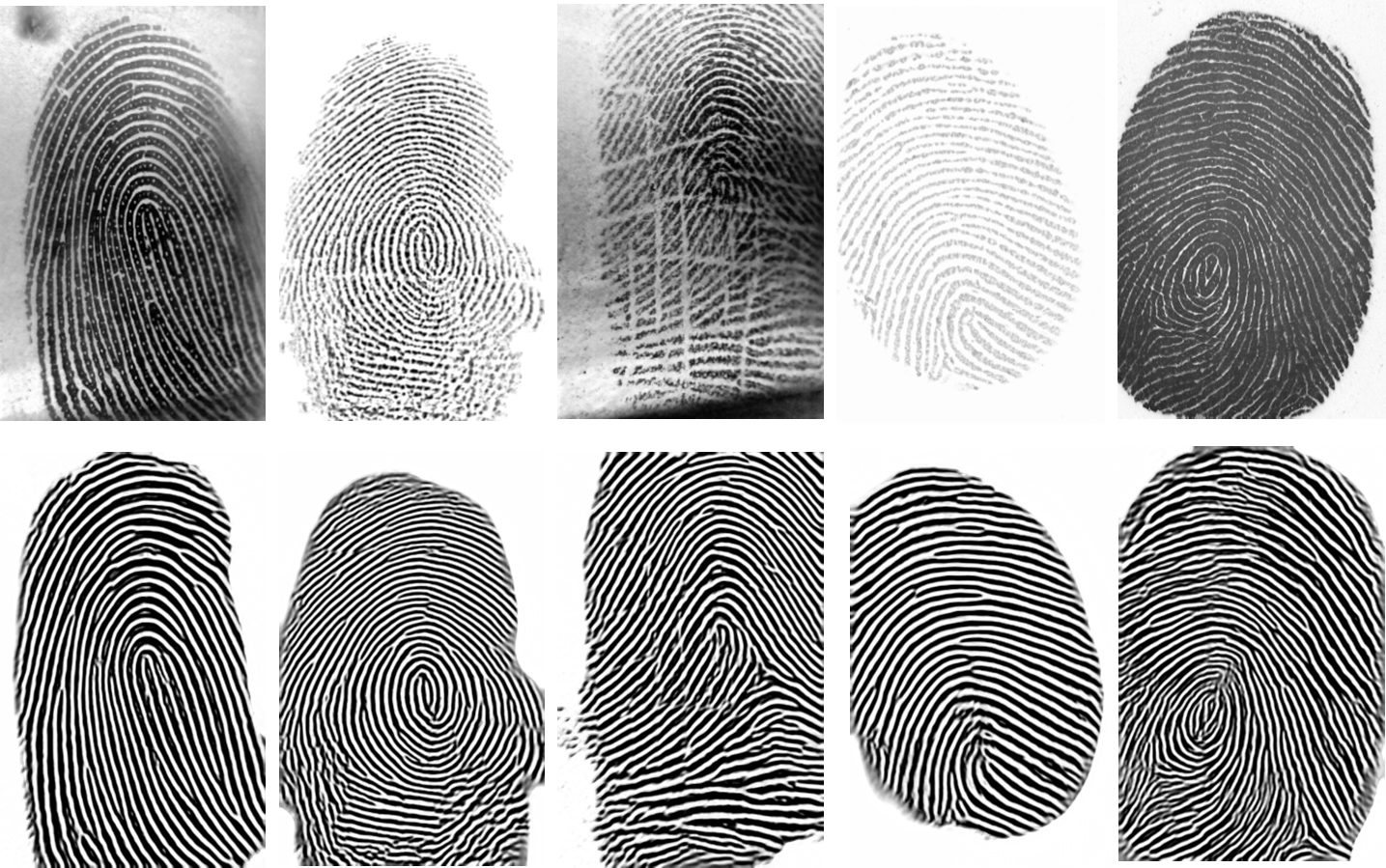}
  \caption{NIST SD302 dataset}
  \label{fig:sd302_results}
\end{subfigure}
\caption{Illustration of enhancement results with our approach on samples from test datasets}
\label{fig:fvc_sd302}
\end{figure*}

\section{\uppercase{Results and Analysis}}
For evaluation, we make use of standard metrics like SSIM, RMSE, and PSNR \cite{adiga2019fpd}. Additionally, we utilize NFIQ2 \cite{851571} package from NIST's NBIS \cite{51496} to measure the quality of fingerprint images. Quality scores can range from 1 to 100. Moreover, we present the average matching scores of genuine pairs on all four subsets of FVC2002 using BOZORTH3 \cite{51496}. SD302 does not contain multiple impressions of a finger, so matching performance can not be obtained.
\\
\\
\noindent \textbf{Ridge Structure Preservation:} In Figure \ref{fig:ssim}, we show a few sample images from our SFinge test set with corresponding ground-truth and enhanced images. We see the SSIM values are higher which suggests that our approach tries to preserve the ridge structure while performing enhancement on degraded input. In addition to this, Table \ref{tab:ablation1} reports SSIM values for various combinations of techniques suggested in this paper. In Figure \ref{fig:fvc_sd302}, we show the results of our approach on both datasets.
\\
\\
\noindent \textbf{Fingerprint Quality Analysis:} We report the average NFIQ2 scores on the test sets FVC 2002 and NIST SD302 in Table \ref{tab:nifq_scores}. From the results, we say that the fingerprint quality improved by a significant amount of 58\% in the case of FVC 2002, whereas it improved by 23\% in the SD302 dataset after enhancing the raw images with our approach. Further, our approach gives comparable results with the previous works. Moreover, we also present the NFIQ2 scores on the SFinGe test set in Table \ref{tab:ablation1} which supports our approach to use WA Block, depthwise separable convolutions, and domain knowledge.
\\
\\
\noindent \textbf{Matching Performance:} We report and compare the average matching scores of genuine pairs of our approach with raw images and previous works in Table \ref{tab:matching}. The results suggest our approach with the inclusion of domain knowledge from the minutia and orientation branch is able to retain the minutiae from the degraded images which increases the matching score and performs well in comparison to earlier works. This suggests the effectiveness of our approach in the pipeline of fingerprint matching. 
\begin{figure}
  \centering \includegraphics[height=4cm, ]{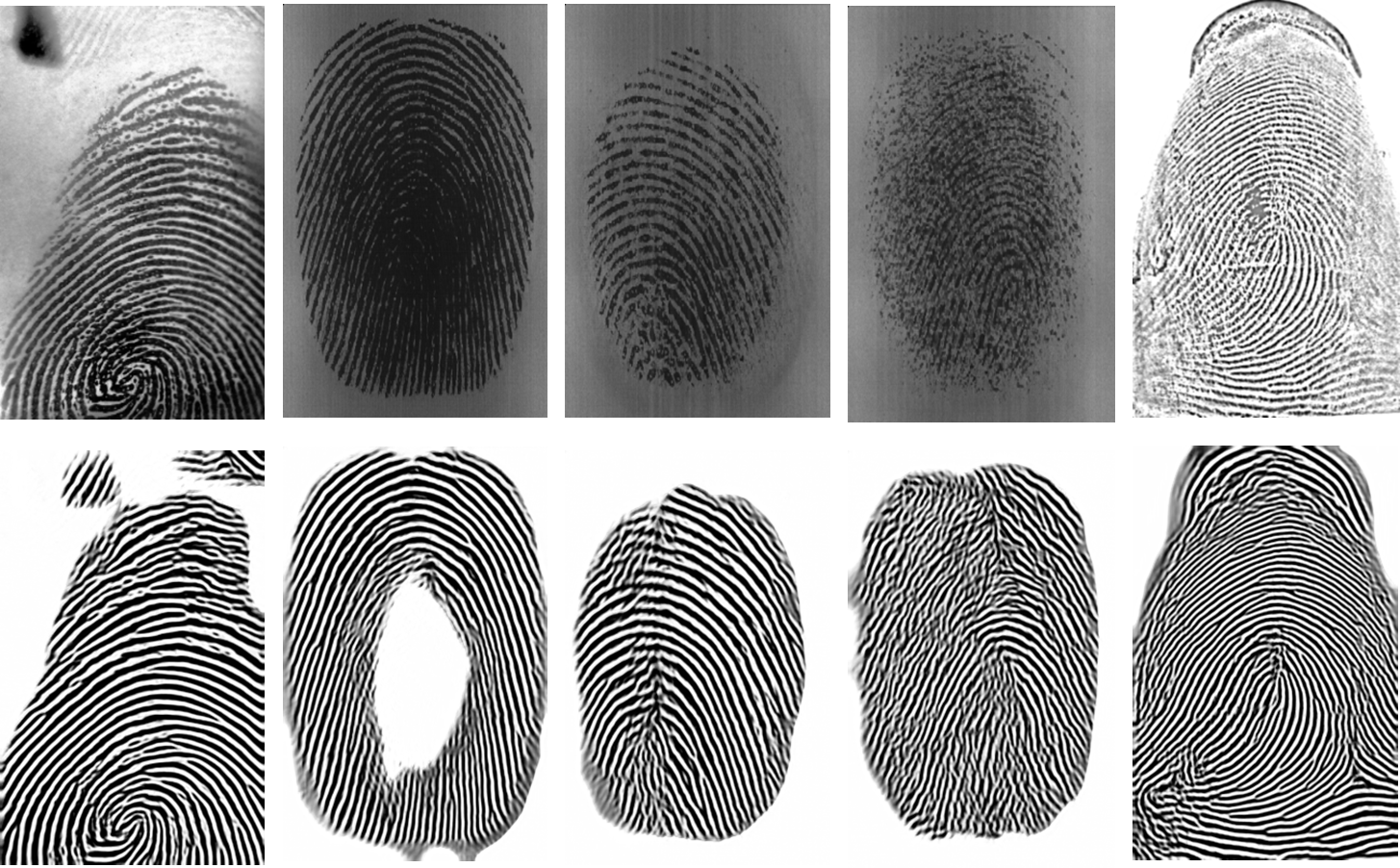}
  \caption{Illustration of failure scenarios observed with our approach. Top row represents degraded input and bottom row denotes the enhanced images with our model.}
  \label{fig:challenges}
\end{figure}
\\
\\
\noindent \textbf{Ablation Study:} We report the evaluation metrics for the techniques discussed in the paper as a part of the ablation study in Table \ref{tab:ablation1}. We show how each discussed method in this paper helps to improve the enhancement further. The depthwise separable convolution do not significantly impact the performance but help reduce the model parameters to a large extent. Multi-task learning with minutia detection and orientation branches gives a clear performance improvement. Moreover, the use of wavelet attention block further improves the SSIM and NFIQ2. Overall, the combination of these techniques results in the best performing model.
\\

% NEW TABLE%
\begin{table}[tb]
% \begin{table*}[]
\small
\centering
\renewcommand{\arraystretch}{1.4} %expand the cells
% \parbox{\textwidth}{\caption{Ablation study: Evaluation performance on SFinGe test dataset with SSIM, MSE, MAE and PSNR metrics with different modifications to U-Net suggested in this paper.}
\caption{Ablation study: Evaluation performance on SFinGe test dataset with different modifications to U-Net suggested in this paper.}
\label{tab:ablation1}
\begin{tabular}{|l|cccc|}
\toprule
\multicolumn{1}{|c|}{\textbf{Approach}} & \textbf{SSIM} & \textbf{RMSE} & \multicolumn{1}{l}{\textbf{PSNR}} & \multicolumn{1}{l|}{\textbf{NFIQ2}} \\ \midrule
\multicolumn{1}{|c|}{Raw Images}    & 0.605 & 117.31 & 6.89 & 36.42 \\ 
\multicolumn{1}{|c|}{($l_2$ loss)}  & 0.863 & 43.67 & 12.78 & 47.72 \\ 
\multicolumn{1}{|c|}{($l_1$ loss)}  & 0.883 & 41.83 & 14.12 & 49.45 \\ 
\multicolumn{1}{|c|}{Depth. Sep. (DS)}                        & 0.890 & 41.25 & 14.17 & 49.78 \\ 
\multicolumn{1}{|c|}{WA Block (WA)}                                & 0.919 & 40.31 & 14.43 & 51.01 \\ 
\multicolumn{1}{|c|}{Minutia (M)}     & 0.934 & 37.92 & 17.15 & 52.86 \\ 
\multicolumn{1}{|c|}{Orientation (O)} & 0.928 & 38.58 & 16.81 & 51.32                            \\ 
\multicolumn{1}{|c|}{M+O}                     & 0.943 & 37.43 & 17.62 & 53.11                      \\ 
\multicolumn{1}{|c|}{M+O+WA}                        & 0.954  & 36.91 & 18.47 & 55.32                       \\ 
\multicolumn{1}{|c|}{M+O+WA+DS} & 0.955  & 36.78 & 18.81 & 55.34                       \\ \bottomrule
\end{tabular}
\end{table}
% NEW TABLE - END%

\begin{table}
\caption{Average NFIQ2 scores of the images from FVC 2002 and NIST SD302 datasets. Higher scores represent higher fingerprint image quality.}
\label{tab:nifq_scores}
\begin{tabular}{@{}|c|c|c|@{}}
\toprule
\textbf{Dataset}             & \textbf{FVC 2002} & \textbf{NIST SD302} \\ \midrule
Raw Images          & 35.10      & 46.97         \\
Joshi et al. & 54.11      & 56.84         \\
Hong et al. & 56.01      & 58.23         \\
Ours                & 56.26      & 58.49         \\ \bottomrule
\end{tabular}
\end{table}

\begin{table}
\caption{Average matching scores from BOZORTH3 on different subsets (DB1, DB2, DB3 and DB4) of FVC 2002 dataset. Higher the scores, better is the approach. Feature extraction was performed using MINDTCT.}
\label{tab:matching}
\begin{tabular}{|c|c|c|c|c|}
\toprule
Database             & DB1 & DB2 & DB3 & DB4 \\ \midrule
Raw Images          & 52.77 & 48.62 & 45.21 & 50.26 \\ Joshi et al. & 71.34 & 71.06 & 67.08 & 68.20 \\ Hong et al. & 73.52 & 72.31 & 69.12 & 70.29 \\
Ours                & 74.01 & 72.61 & 69.10 & 71.08 \\ \bottomrule
\end{tabular}
\end{table}

\noindent \textbf{Observed Challenges:} We observed that our approach Finger-UNet performs well from our experimental results. However, we also observed scenarios where our approach did not perform well. A few of these cases are demonstrated in Figure \ref{fig:challenges}. We saw that in cases where the input has severe artifacts, the model mistakes it for a portion of the fingerprint and tries to enhance it. Moreover, in cases where the input is too dark or too light to correctly figure out the ridge structure, the model fails to predict good fingerprints in those areas. In addition, if the input contains a nail, it fails to discriminate the nail from the fingerprint. We believe these issues arise as the network has not seen such data during training, as the SFinge dataset does not contain samples with nails or artifacts.

\section{\uppercase{Conclusion}}
\label{sec:conclusion}
In this paper, we modified vanilla U-Net, combining multiple techniques to improve the fingerprint quality with synthetic data from SFinGe. We evaluated our model on two public fingerprint datasets FVC 2002 and NIST SD302. The network is robust enough to recover fingerprints even with various degrees of degradation. From the experimental results, we say that $l_1$ loss performed well for this task, along with domain knowledge from minutia and orientation branches, which improved performance above baselines. We also discussed the challenging cases in our experiments and the possible solutions. Moreover, using the wavelet attention block helped improve the performance.
In future works, we plan to explore and design new quality metrics better suited for fingerprint enhancement along with other deep architectures.

\bibliographystyle{apalike}
{\small
\bibliography{example}}

% \section*{\uppercase{Appendix}}

% If any, the appendix should appear directly after the
% references without numbering, and not on a new page. To do so please use the following command:
% \textit{$\backslash$section*\{APPENDIX\}}

\end{document}